\documentclass[11pt,a4paper]{article}
\usepackage[hyperref]{acl2021}
\usepackage{times}
\usepackage{latexsym}

\usepackage{microtype}
\usepackage{graphicx}

\aclfinalcopy 

\title{Polite Emotional Dialogue Acts for Conversational Analysis \\in Daily Dialog Data}

\author{Chandrakant Bothe \\
	\url{https://bothe.in}\\
	PhD, University of Hamburg\\
	Chief AI Officer, Foviatech GmbH\\
	bothe.chandrakant@gmail.com\\}

\date{}

\begin{document}
	\maketitle
	\begin{abstract}
		Many socio-linguistic cues are used in the conversational analysis, such as emotion, sentiment, and dialogue acts.
		One of the fundamental social cues is politeness, which linguistically possesses properties useful in conversational analysis.
		This short article presents some of the brief findings of polite emotional dialogue acts, where we can correlate the relational bonds between these socio-linguistics cues.
		We found that the utterances with emotion classes Anger and Disgust are more likely to be impolite while Happiness and Sadness to be polite.
		Similar phenomenon occurs with dialogue acts, Inform and Commissive contain many polite utterances than Question and Directive.
		Finally, we will conclude on the future work of these findings.
	\end{abstract}

	\section{Introduction}
	
	Conversational analysis can potentially be enhanced with the help of politeness socio-linguistic cues \cite{P13dan2013compPoliteness} along with other linguistic cues.
	It is also highlighted in the concluding remarks of the PhD thesis Conversational Language Learning for Human-Robot Interaction \cite{bothe2020conversational}.
	This article\footnote{This article is written and published by the author in his own interest and capacity, without any responsibility and conflict of interest of mentioned affiliation.} focuses on a brief extension of the experiments provided in the mentioned thesis, primarily by adding linguistic politeness cues to conversational analysis along with emotion states and dialogue acts.

\begin{figure}[t!]
		\begin{center}
			\includegraphics[width=\linewidth]{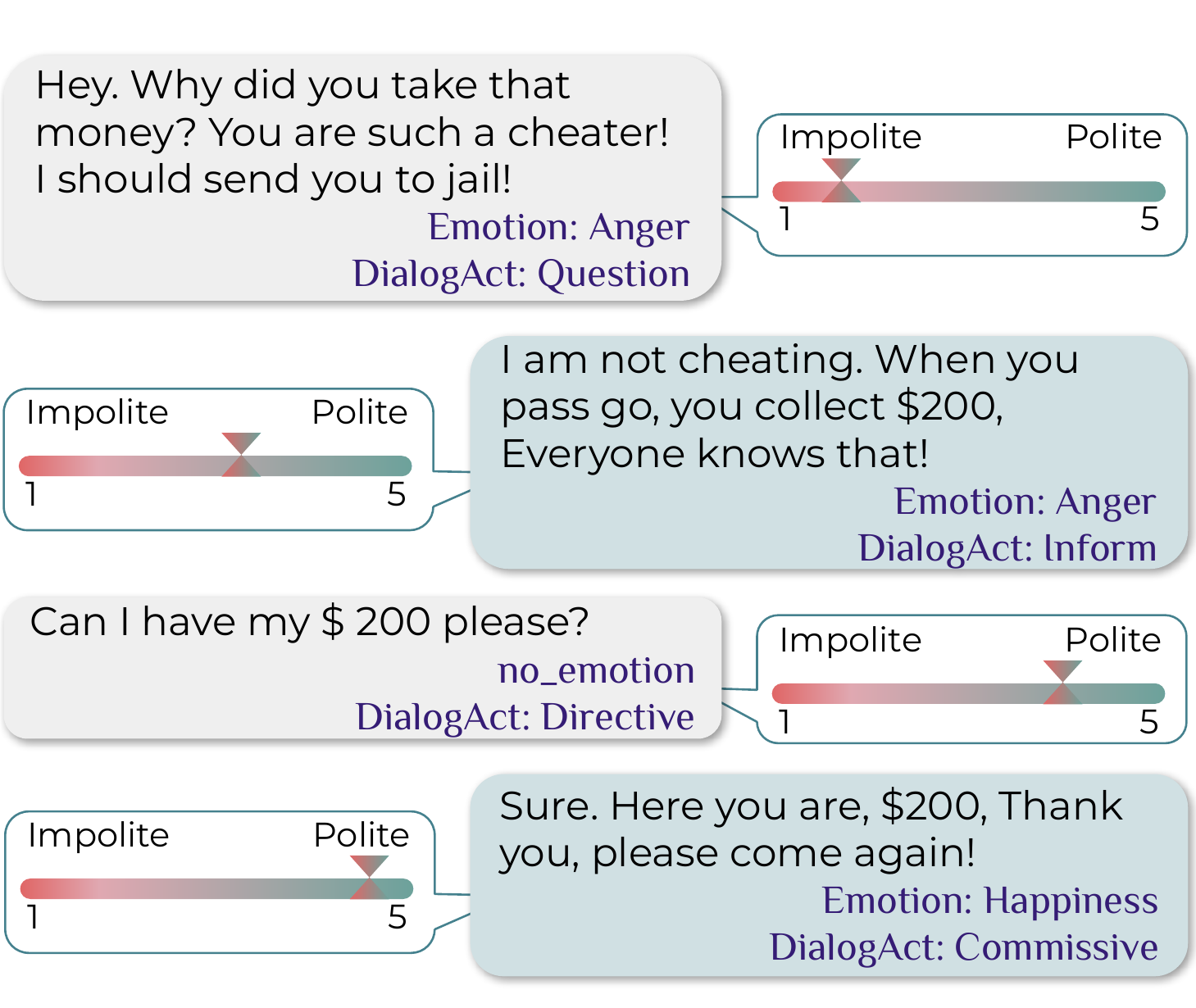} 
			\caption{Polite Emotional Dialogue Acts labels from the DailyDialog dataset, 
				Example shows Anger with very Impolite and Happiness with very Polite}
			\label{ang-happ-pols}
		\end{center}
	\end{figure}

	In this brief experiment, we will explore DailyDialog dataset \cite{dailydialog2017}, a multi-turn dialogue dataset, and it is already annotated with emotion and dialogue act labels.
	We will find how this dialogue dataset is annotated with a pre-trained politeness recognizer \cite{jiajun2021conversation} that provides values from 1 to 5 on a politeness scale. 
	We discover that the utterances with certain emotion classes are more polite or impolite than others.
	At least most of the time, it seems natural to use polite utterances in a happy emotional state, whereas to use impolite utterances in an angry state.
	This phenomenon is precisely the motivation behind this experiment; find some utterance examples given in Figure~\ref{ang-happ-pols}.
	The conversation in this example shows how an impolite utterance is used in the Anger emotion state and then ends with a polite utterance in the Happiness emotion state.
	We find out such phenomenon statistically occurring quite often in the given dataset, and the results are also available in the GitHub repository bothe/politeEDAs\footnote{\url{https://github.com/bothe/politeEDAs}}.

	\begin{table*}[t!]
		\centering
		\begin{tabular}{lrrrrrrr}
			\hline      \multicolumn{4}{c}{Emotions} & \multicolumn{4}{c}{Dialogue Acts} \\ 
			\hline      & Utt   & \% total & \% Emo & Inform & Question & Directive & Commissive \\ \hline
			Anger       & 1022  & 1.00  &  5.87 &   615 &   174 &   132 &  101 \\
			Disgust     & 353   & 0.34  &  2.03 &   291 &    30 &    18 &   14 \\
			Fear        & 174   & 0.17  &  1.00 &   106 &    27 &    21 &   20 \\
			Happiness   & 12885 & 12.61 & 74.02 &  7830 &  2158 &  1476 & 1421 \\
			Sadness     & 1150  & 1.13  &  6.61 &   809 &   190 &    95 &   56 \\
			Surprise    & 1823  & 1.78  & 10.47 &  1122 &   565 &    72 &   64 \\
			no\_emotion & 85572 & 83.78 &    -- & 36316 & 26549 & 15542 & 7165 \\
			\hline
			Total       &102979 &  --   &    -- & 47089 & 29693 & 17356 & 8841 \\
			\hline
		\end{tabular}
		\caption{\label{emo-data-table}Statistics of utterances with Emotion and Dialogue Act classes in the DailyDialog dataset}
	\end{table*}

	\section{Related Work}
	
	Perceiving emotions in conversation provides affective information of the conversation partners; similarly, perceiving politeness in conversation provides cues about their social manners.
	The emotional states and dialogue acts have unique relations which is presented by \citeauthor{bothe2019enriching} in Emotional Dialogue Acts \cite{labothe2017WASSA2017,bothe2018discourse,bothe2019enriching}. 
	For example, Accept/Agree and Thanking dialogue acts often occur with the Joy emotion, Apology with Sadness, Reject with Anger, and Acknowledgements with Surprise.
	Similarly, politeness and emotion together are evidently discussed a lot in the literature as some of the most critical social cues \cite{Langlotz2017polemo,renner2020directness,Culpeper2021politeness}.
	Some computational linguistic study shows how machines learn politeness, for example, \textit{please} and \textit{could you} words signals on the heatmaps of sentences \cite{aubakirova2016interpoliteness}.

	In fact, \citet{brown1987politeness} mentions that the display of emotions or lack of control of emotions as positive politeness strategies or potentially face-threatening acts.
	Thus this study fosters analysis of linguistic politeness by understanding how ‘appropriate levels of affect’ are conveyed in the conversational interaction \cite{Langlotz2017polemo}.
	In human-robot interaction, politeness as a social cue plays a vital role to drive engaging social interaction with robots \cite{steinfeld2006common,srinivasan2016help,bothe2018towards}.
	However, this article explores relations between the socio-linguistic cues, finding out how polite the utterances are against their respective emotion and dialogue act classes.

	\section{Approach}
	
	Our goal is to analyze the socio-linguistic cues to find meaningful relations between them, for which we are going to perform the tasks in the following order:
	\begin{itemize}
		\item explore the dataset, DailyDialog, which is already annotated with emotion and dialogue act labels,
		\item annotate the utterances with politeness values using politeness analyzer, more precisely each utterance with a degree of politeness,
		\item analyze the annotated utterances with politeness against the emotion and dialogue act classes in the dataset, and
		\item conclude with the results and findings and provides future conversational analysis work to extend this experiment.
	\end{itemize}

	\section{Resources}
	
	\subsection{Dataset: DailyDialog}
	
	As the name suggests, the DailyDialog dataset contains conversation topics of daily life ranging from ordinary life to financial topics.
	It contains bi-turn dialogue flow such as Question-Inform and Directive-Commissive; thus, the dataset is annotated with those four fundamental dialogue acts to follow unique multi-turn dialogue flow patterns \cite{dailydialog2017}. 
	This dataset is manually labelled with six emotion classes and no\_emotion class.
	The statistics of the dataset is presented in Table~\ref{emo-data-table}, we can see that the no\_emotion class dominates the total number of utterances.
	However, Happiness dominates within emotion classes, whereas Fear contains the least number of utterances.
	Table~\ref{emo-data-table} also presents the number of utterances for the dialogue act classes in the dataset for their corresponding emotion classes.

	\begin{figure*}[t!]
		\begin{center}
			\includegraphics[width=\textwidth]{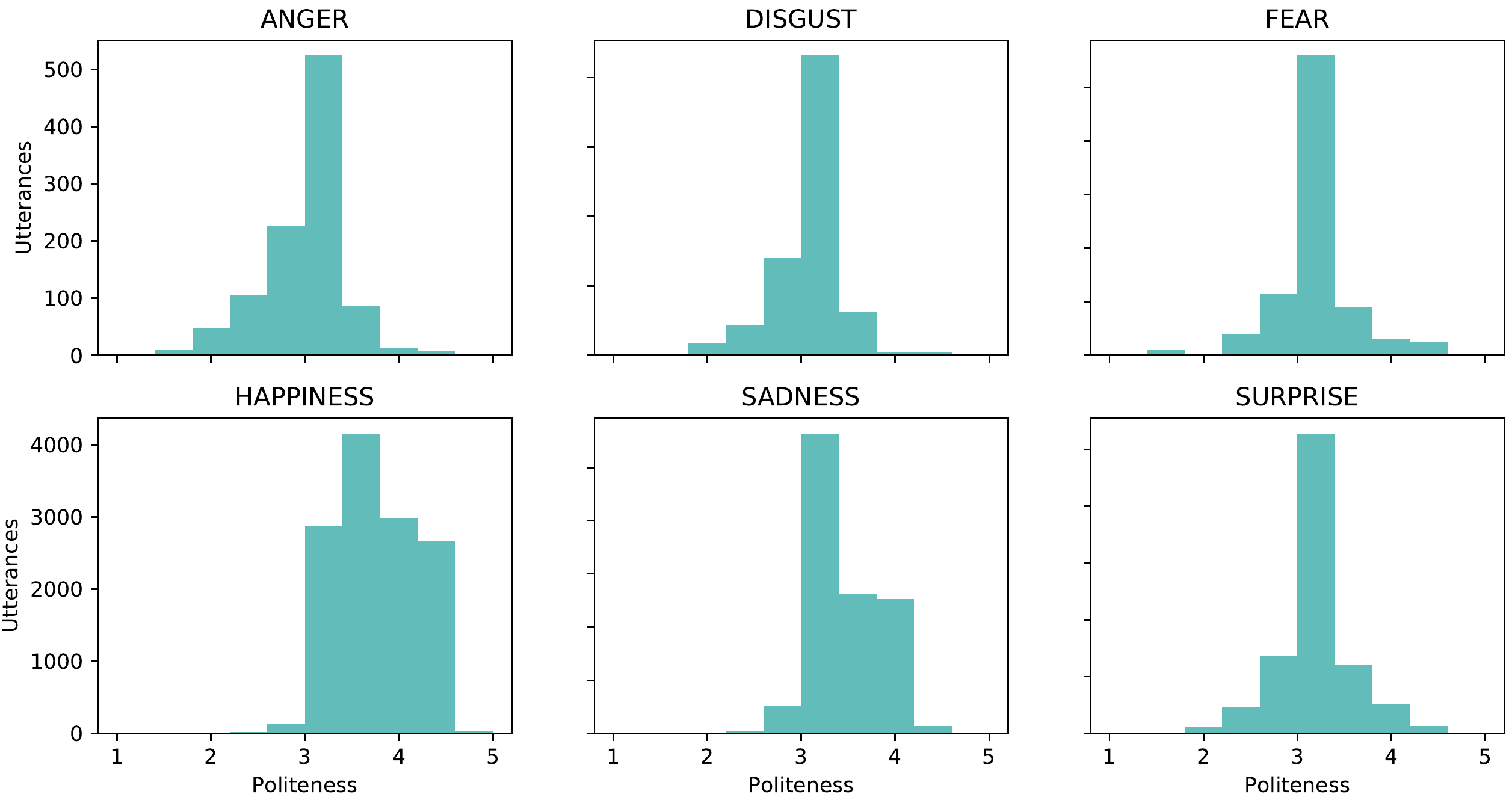} 
			\caption{Politeness histograms for the Emotion categories in the DailyDialog dataset}
			\label{pol_histo_emo}
		\end{center}
	\end{figure*}

	\subsection{Politeness Analyzer}
	
	We will use a politeness analyzer from the recent work of \citet{jiajun2021conversation}, where they combine two datasets from \citet{P13dan2013compPoliteness} and \citet{wang2018convosupport}.
	The politeness regressor model is obtained on these datasets by pre-training a BERT-based model \cite{devlin2018bert} on Reddit data using masked language modelling. 
	The final model is obtained at an average Pearson of 0.66 with human judgments from both the datasets and made available online\footnote{\url{https://github.com/wujunjie1998/Politenessr/}} by authors.
	The model provides a degree of politeness on the scale in [1, 5], where around 3 indicates neutral.

	\section{Results and Analysis}
	
	We annotate all utterances in the DailyDialog dataset for politeness values in the range [1, 5] using the politeness analyzer mentioned above.
	Then we plot histograms of the politeness values of the utterances against the emotion categories shown in Figure~\ref{pol_histo_emo}.
	As we can see in these histograms, utterances in the emotion classes Anger and Disgust are neutral and impolite on the politeness scale. 
	On the other hand, many utterances in the Happiness and Sadness emotion classes are polite. 
	Fear and Surprise classes are mostly neutral on the politeness scale than other emotion classes.

	\begin{figure}[b!]
		\begin{center}
			\includegraphics[width=0.5\textwidth]{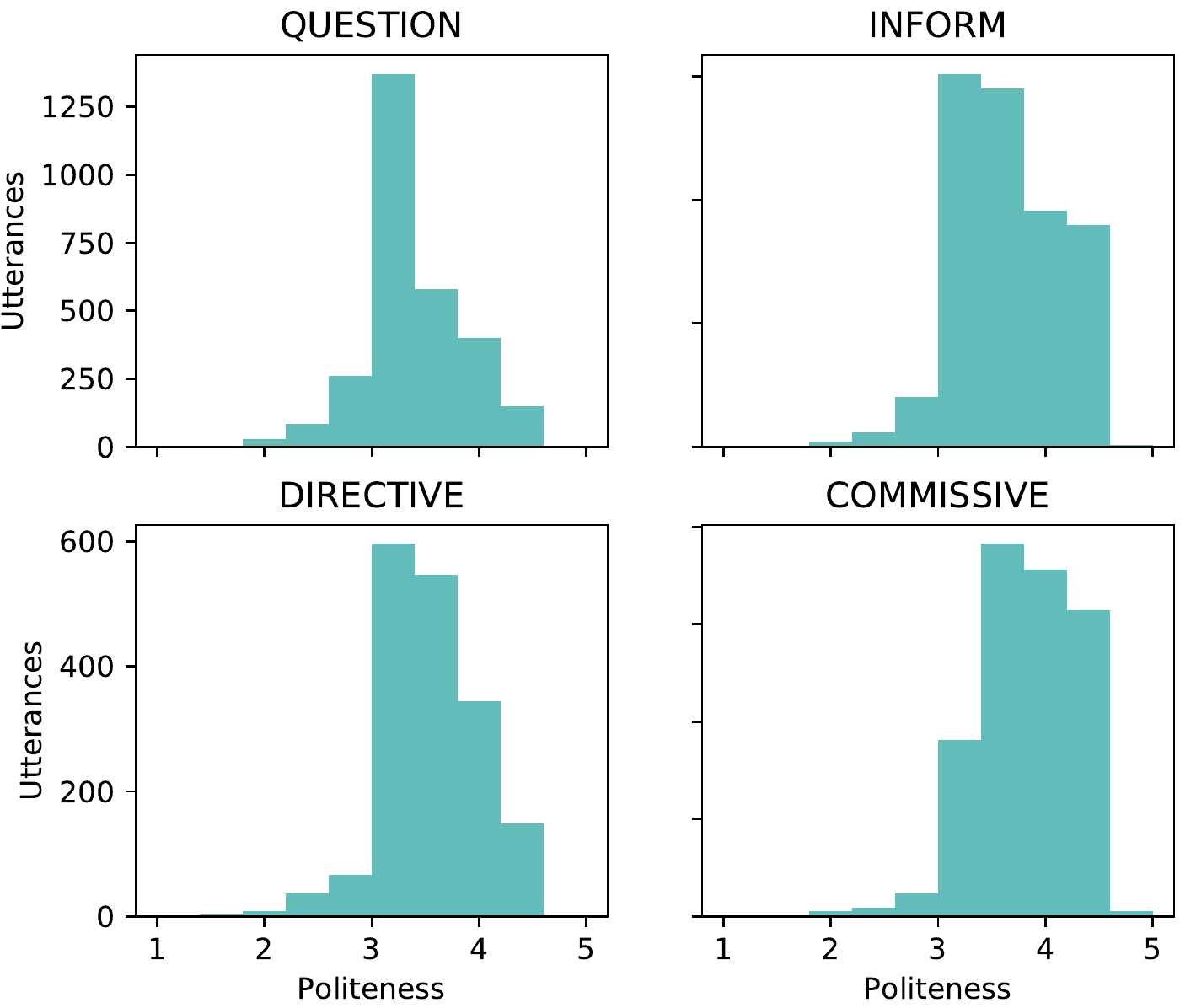}
			\caption{Politeness Histograms for the Dialogue Acts in the DailyDialog dataset}
			\label{pol_histo_da}
		\end{center}
	\end{figure}

	\begin{figure}[b!]
		\begin{center}
			\includegraphics[width=0.48\textwidth]{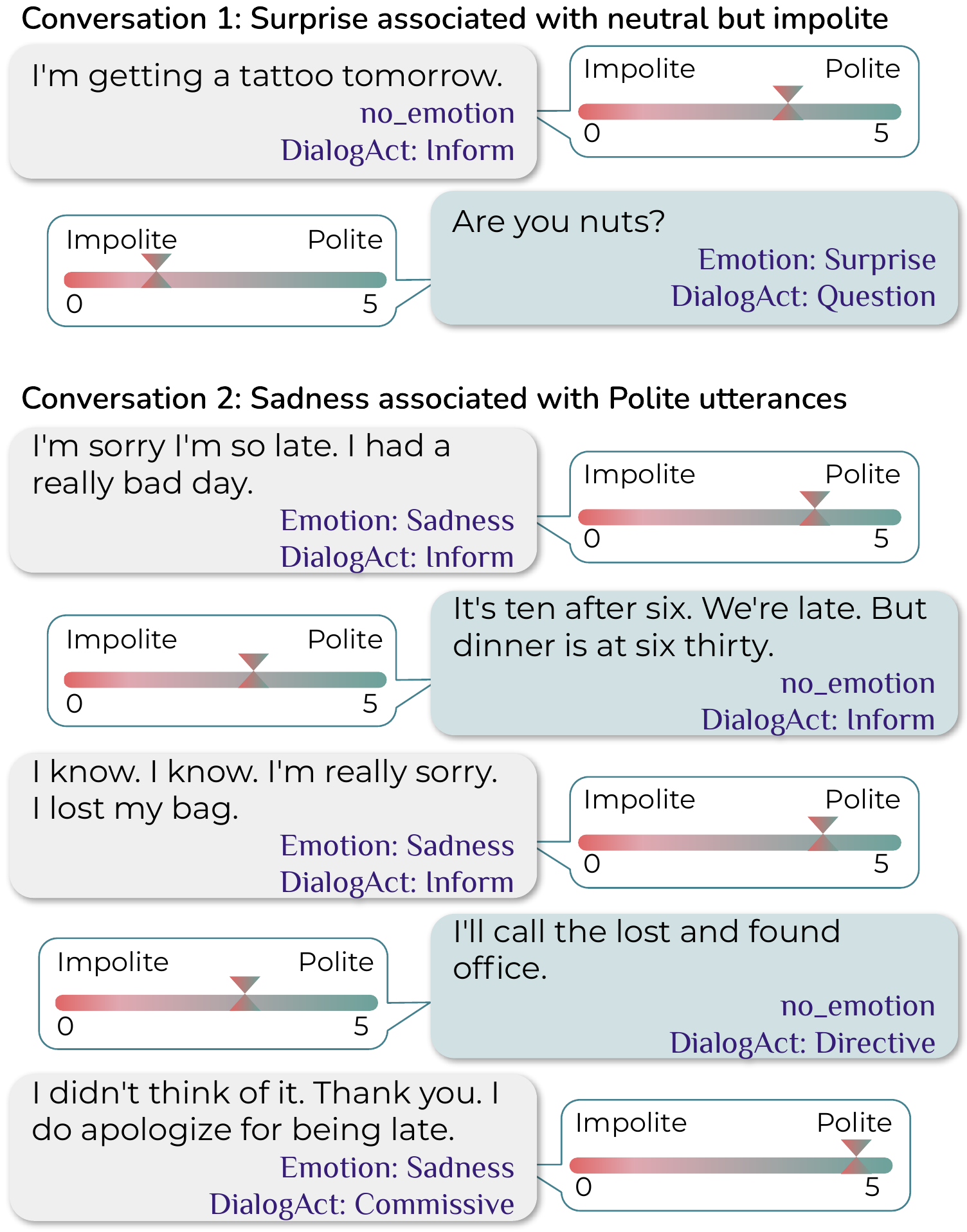} 
			\caption{Polite Emotional Dialogue Act labels from the DailyDialog dataset focused on Surprise and Sadness emotion utterances, the utterance with Surprise shows an immediate reaction which becomes impolite,
				interestingly most of the Sadness utterances are polite}
			\label{emotion_pol_impol_dial}
		\end{center}
	\end{figure}

	We also plot the histograms of the politeness values of the utterances against the dialogue act classes shown in Figure~\ref{pol_histo_da}.
	We can see that most of the utterances are neutral or polite on the politeness scale for all the dialogue acts.
	However, we find that the Inform and Commissive dialogue act classes contain many polite utterances compared to Question and Directive dialogue acts.
	Interestingly, we discovered that speakers often tend to use polite utterances when answering, thanking, and agreeing (and the utterances that come under Inform and Commissive).
	Whereas when asking, guiding, directing, ordering (under Question and Directive), the utterances are primarily neutral or polite.
	
	\begin{table*}[t!]
		\centering
		\begin{tabular}{llll}
			\hline DoP & Dialogu Act & Emotion & Utterances  \\ \hline 
			\multicolumn{3}{l}{\underline{Polite utterances}} \\
			4.63 & INFORM & HAPPINESS & OK! Thank you very much! \\
			4.63 & INFORM & HAPPINESS & Sure, that would be great! Thank you!\\
			4.63 & COMMISSIVE & HAPPINESS & That would be great! Thanks a lot!\\
			4.62 & INFORM & HAPPINESS & Thank you, thank you, thanks again.\\
			4.62 & INFORM & HAPPINESS & This is exciting! Thank you so much!\\
			\hline
			\multicolumn{3}{l}{\underline{Neutral utterances}} \\
			3.34 & INFORM & HAPPINESS & I went to the tutoring service centre on campus today. \\
			3.34 & INFORM & HAPPINESS & I never knew there were so many fun things to do on a farm. \\
			3.34 & INFORM & no\_emotion & Just a minute. It's ten to nine by my watch. \\
			3.34 & INFORM & no\_emotion & You may check out books or videos. \\
			3.34 & QUESTION &  no\_emotion & That’s a small fee? \\
			\hline
			\multicolumn{3}{l}{\underline{Impolite utterances}} \\
			1.59 & INFORM & DISGUST & Don’t dress like that. You’ll make fool yourself. \\ 
			1.51 & COMMISSIVE & ANGER & Make it work, Geoff. You would say that, wouldn’t you. \\
			1.47 & DIRECTIVE & no\_emotion & Get up, you lazybones! \\
			1.46 & DIRECTIVE & SURPRISE & You idiot! Don’t say that! Do you want this job, or not? \\
			1.37 & DIRECTIVE & ANGER & Get out of my store, you jerk! \\
			\hline
		\end{tabular}
		\caption{\label{anno-pols-table} Examples from the DailyDialog dataset showing 5 utterances extreme DoP (degree of politeness)}
	\end{table*}

	Table~\ref{anno-pols-table} presents top polite, neutral and impolite utterances from the DailyDialog dataset annotated with the politeness analyzer.
	There are five examples from the top 10 in each of the range of extreme values (only 5 examples are shown out of the top 10 to eliminate similar or repeated utterances).
	We can see that most of the utterances with ``Thank you" phrases are recognized with a significantly higher degree of politeness. 
	In contrast, most Anger and Disgust utterances are recognized with a significantly lower degree of politeness.
	Figure~\ref{emotion_pol_impol_dial} provides two conversation examples of Surprise and Sadness emotion utterances.

	\section{Conclusion}
	
	Recognizing politeness in conversation is an essential aspect of conversational analysis, and we discover how politeness flows in a dialogue. 
	Moreover, socio-linguistics features relatedness provide an additional dimension for behavioural analysis of conversation partners and their use in the virtual assistant and human-robot interaction fields.
	
	In this short paper, we discover how politeness is related to emotions and dialogue acts in the given dataset.
	Specifically, we discover that the utterances are mostly polite in Happiness and Sadness emotion classes and Inform and Commissive dialogue acts.  
	Similarly, the utterances are primarily neutral or impolite in Anger and Disgust emotion classes.
	We find that the utterances in the Question and Directive dialogue acts are mostly neutral; however, many are also polite.

	The continuing work under this experiment is planned to extend to other dialogue datasets to prove the discovered phenomenon.
	The politeness analyzer used in this experiment provided fair annotations for the degree of politeness linguistically.
	However, the next step would be to use more than one analyzer for more robust detection of politeness or annotate the utterances manually. 
	Future work could also be extended to learn the social behaviours using analyzed linguistic cues with the help of deep learning techniques.


	\bibliographystyle{acl_natbib}
	\bibliography{acl2021}
	
	
\end{document}